# Evaluating the Impact of Different Quantum Kernels on the Classification Performance of Support Vector Machine Algorithm: A Medical Dataset Application

Emine AKPINAR*
Department of Physics, Yildiz Technical University, Istanbul, Turkey

Sardar M. N. ISLAM
Applied Informatics Program, ISILC, Victoria University, Melbourne, Australia

Murat ODUNCUOGLU
Department of Physics, Yildiz Technical University, Istanbul, Turkey

*Corresponding author(s). E-mail(s): emineakpinar28@gmail.com

**Abstract**— The support vector machine algorithm with a quantum kernel estimator (QSVM-Kernel), as a leading example of a quantum machine learning technique, has undergone significant advancements. Nevertheless, its integration with classical data presents unique challenges. While quantum computers primarily interact with data in quantum states, embedding classical data into quantum states using feature mapping techniques is essential for leveraging quantum algorithms. Despite the recognized importance of feature mapping, its specific impact on data classification outcomes remains largely unexplored. This study addresses this gap by comprehensively assessing the effects of various feature mapping methods on classification results, taking medical data analysis as a case study. In this study, the QSVM-Kernel method was applied to classification problems in two different and publicly available medical datasets, namely, the Wisconsin Breast Cancer (original) and The Cancer Genome Atlas (TCGA) Glioma datasets. In the QSVM-Kernel algorithm, quantum kernel matrices obtained from 9 different quantum feature maps were used. Thus, the effects of these quantum feature maps on the classification results of the QSVM-Kernel algorithm were examined in terms of both classifier performance and total execution time. As a result, in the Wisconsin Breast Cancer (original) and TCGA Glioma datasets, when $R_x$ and $R_y$ rotational gates were used, respectively, as feature maps in the QSVM-Kernel algorithm, the best classification performances were achieved both in terms of classification performance and total execution time. The contributions of this study are that (1) it highlights the significant impact of feature mapping techniques on medical data classification outcomes using the QSVM-Kernel algorithm, and (2) it also guides undertaking research for improved QSVM classification performance.

## I. Introduction

Quantum machine learning (QML) is a rapidly evolving research field where the power of quantum computing is applied to machine learning (ML) tasks to pursue computational advantages over data that, in the current context, is becoming increasingly difficult to manage, process and draw meaningful conclusions from [1, 2]. Quantum computers possess parallel processing power due to the superposition and entanglement properties of quantum particles. In this case, the primary objective in QML research is to utilize quantum mechanical concepts such as superposition and quantum entanglement to enhance the performance of ML techniques. Nevertheless, despite the great potential of quantum computing, current noisy intermediate-scale quantum (NISQ) computers have limited qubits and low circuit depths [3]. Furthermore, these computers are highly susceptible to their ambient surroundings and are at risk of losing their quantum state because of quantum decoherence [4]. Despite these hardware limitations, there are already quantum versions of some ML models in the literature, such as quantum principal component analysis [5], a popular approach for dimensionality reduction, quantum Gaussian mixture models [6], another renowned approach for estimating clustering and density, and a support-vector machine algorithm with a quantum kernel estimator (QSVM-Kernel), which is basically applied for linear and nonlinear classification problems [7].

The QSVM-Kernel algorithm is a quantum adaptation of the classical support vector machine (SVM) algorithm that leverages the capabilities of quantum mechanics, such as superposition and entanglement, to augment the kernel calculation process [7-9]. The motivation for using quantum kernels is based on their ability to partition the data/input space more discriminatively than classical calculations can. The quantum kernel method $\left(k(\vec{x_i}, \vec{x_j}) = \left|\langle U_\Phi(\vec{x_i}) | U_\Phi(\vec{x_j}) \rangle\right|^2 = \left|\langle \Phi(\vec{x_i}) | \Phi(\vec{x_j}) \rangle\right|^2\right)$ employs the inner product of quantum feature maps to encode data from classical data space to corresponding qubit states in an exponentially large target space, namely, the Hilbert space [10]. In other words, the data are nonlinearly mapped to a high-dimensional feature space, where a hyperplane is constructed to separate the labeled samples. To outperform classical approaches, quantum feature maps should be complex enough to be difficult to simulate on classical computers and at the same time ideally suited for implementation on NISQ computers [7, 11].

Feature mapping techniques in kernel methods are pivotal for pattern recognition and feature detection, as they aid in identifying labeled patterns in datasets where they may be difficult to distinguish in their original space [12]. However, as the feature space grows and the computational cost of estimating kernel functions increases, traditional methods face limitations in successfully addressing such classification problems. To overcome this challenge, data are mapped to quantum state in the high-dimensional Hilbert space using quantum feature mapping methods, rather than the classical data space. In other words, this process entails nonlinearly mapping classical data ($\vec{x}$) to a quantum state ($\vec{x} \rightarrow |\langle \Phi(\vec{x_i}) | \Phi(\vec{x_j}) \rangle|$). In QML, several different encoding (embedding) procedures have been proposed for encoding data quantum mechanically, such as angle embedding [10, 13], amplitude embedding [14, 15], Pauli feature maps [7, 9, 16], parametrized quantum circuits [17-19].

Here, we report an experimental implementation of the QSVM-Kernel algorithm on the classification performance of two different medical datasets as case studies.

In recent years, with the proliferation of comprehensive medical public databases integrating electronic health records, billing data, disease registries, medical imaging, biomarker data, wearable technologies, and health applications, new strategies and methods have been designed and developed to expedite complex data analyses, manage large datasets, and derive meaningful insights [20, 21]. Furthermore, with these open databases, numerous powerful artificial intelligence (AI) decision support systems, notably machine learning (ML) models, have been developed to address significant problems in medical data, including tumor classification [22-28], noninvasive determination of genomic biomarkers for diseases [29], prediction of patient survival outcomes [30], and prediction of the onset of various pathological conditions [31]. In particular, medical data classification is a crucial application of ML in the healthcare domain. Indeed, some of these techniques have demonstrated applicability in computer-aided detection systems.

However, as highlighted in many studies in the literature, classical ML techniques have limitations in learning correlations and most complex patterns in medical datasets [32]. The noise in the data, feature dimensionality, and limited nature of the classical vector space are some of the challenges encountered [33]. Moreover, the need for higher computational power increases as the size of medical datasets grows, posing another significant problem [34]. Recent research has shown that in addition to classical ML techniques in healthcare, the utilization of quantum machine learning (QML) technologies not only addresses these challenges but also accelerates complex data analyses and enables more efficient processing of large datasets [35].

This study provides a comprehensive analysis by employing the Scikit-learn SVM on two distinct medical datasets, namely, the Wisconsin Breast Cancer (original) and The Cancer Genome Atlas (TCGA) Glioma datasets, for binary classification tasks. The SVM leverages quantum kernel methods obtained through quantum feature mapping techniques implemented in the PennyLane software framework. These quantum kernels are used as hyperparameters, and the transformed training set is accepted as input data. The SVM subsequently trains on these data in a classical manner, ultimately yielding a separating hyperplane as a result.

The main objective of the article is to investigate the impact of nine different quantum feature mapping techniques, namely, angle encoding ($R_x$, $R_y$, and $R_z$), amplitude encoding, Pauli feature maps (ZFeatureMap, ZZFeatureMap), and parametrized quantum circuits ($R_x$ and CX, $R_y$ and CY, $R_z$ and CZ), created using the PennyLane software framework, on the classification results of two different medical datasets as case studies.

**Contributions of the Study:** This study shows the impact of different quantum kernels on the classification performance of the QSVM. The significance of kernels on the outcome of classification and execution time of the QSVM algorithm has been highlighted. Therefore, this study has made the following contributions: it shows that quantum kernels can improve the performance of SVMs, and (2) it guides undertaking research for improved QSVM classification performance, especially in medical data analysis.

The remainder of this work is structured as follows:

**Section 2** provides detailed information on the medical datasets used in our study, the preprocessing steps, and the QSVM-Kernel algorithm. **Section 3** presents the results of the impact of different quantum feature maps on the classification outcomes of the QSVM-Kernel algorithm, evaluated in terms of both classifier performance and overall execution time. **Section 4** summarizes the main findings of our study and underscores its contribution to the literature.

## II. MATERIALS AND METHODS

### A. Data and Preprocessing

In this study, two different widely used and publicly available medical datasets were utilized for empirical analysis. Both datasets contain multiple variables and present a classification problem that aims to classify examples into different classes using these variables. The first dataset is the Wisconsin Breast Cancer (original) dataset from the UCI Machine Learning Repository, and the second dataset is The Cancer Genome Atlas (TCGA) dataset, which includes molecular features (or markers) for low-grade gliomas (LGGs) and high-grade gliomas (HGGs) and clinical information for patients. The Wisconsin Breast Cancer dataset has 699 patients (Benign: 458, Malignant: 241), and the dataset comprises 9 integer-valued attributes related to benign and malignant tumors. The Wisconsin Breast Cancer (original) dataset can be accessed at [36]. The original TCGA dataset contains molecular features and clinical information for LGGs and HGGs and consisted of 862 patients and 23 features. Of the 862 patients, 23 patients with missing clinical information were excluded from the dataset. The remaining dataset contain information for a total of 839 patients, with 487 belonging to LGGs and the remaining 352 belonging to HGGs. Additionally, out of the 23 features, 3 pertained to the clinical characteristics of the patients, while the remaining 20 represent the most frequently mutated molecular features. Among these 23 features, one of them is an integer-valued attribute, and the others are categorical. The TCGA research network dataset can be accessed at [37]. In the TCGA dataset, these values were converted into numerical values using the one-hot encoding method after identifying the categorical values.

In the data preprocessing step, the following procedures were applied separately to both datasets:

*a) Normalization of the data:*

In QML studies, standardizing the data to a common range is important for maintaining the coherence of quantum states and reducing the impact of quantum noise. This study employed the min–max normalization technique to rescale the numerical data features to a specific range after converting categorical values into numerical ones. Min–max normalization performs a linear transformation on the original data while preserving the relationships between the data values. Min–max normalization assigns data features to a certain minimum and maximum range. This method is usually applied using the following formula:

$$x_{normalize} = \frac{(x - \min(x))}{(\max(x) - \min(x))} \quad (1)$$

Here, $x_{normalize}$ represents the normalized value, $x$ represents the original value, $\min(x)$, is the minimum value, and $\max(x)$ is the maximum value of the data feature.

*b) Dimension reduction with principal component analysis (PCA):*

Due to the limited number of qubits and low circuit depths inherent in NISQ computers, it is necessary to reduce the number of features to a certain extent before they are mapped to quantum space. This necessity is crucial for enhancing computational efficiency because minimizing the feature set assists in better management of the constraints associated with quantum computing power. This process can essentially be achieved in two different ways.

The first approach entails the utilization of feature selection techniques aimed at identifying input variables with the strongest correlations to the output variables, thus selecting the most important features within the dataset. In our previous work, we employed an ensemble feature selection method in the classical computation part of our hybrid model to identify the most significant or informative features in the TCGA dataset. Detailed information regarding the conducted feature selection process can be found in the [38].

The second approach, known as dimensionality reduction, aims to create a lower-dimensional subspace that captures relationships among features rather than directly selecting features. In this study, we employed the dimensionality reduction method principal component analysis (PCA) to mitigate computational costs in both medical datasets. PCA is employed to extract principal components and project input data onto lower dimensions. The objective is to identify new orthogonal sets of principal components representing directions of maximum variance in the data [39]. The input data are obtained from the principal component with the highest variance. While this method reduces computational costs and expedites the classification process, it does not guarantee the selection of the most discriminative components. In our study, we compressed both datasets using PCA and utilized the first 5 dimensions to represent the raw data.

*B. Support Vector Machine Algorithm with a Quantum Kernel Estimator (QSVM-Kernel)*

This section provides a detailed explanation of the Quantum Support Vector Machine (QSVM-Kernel), the quantum version of the classical SVM model with a quantum kernel estimator.

*a) Support Vector Machine (SVM)*

SVM is a supervised machine learning algorithm with a robust theoretical foundation based on statistical learning theory. Supervised learning is the process of determining a relationship $f(x)$ by using a training dataset $S = \{(x_1, y_1), \ldots, (x_2, y_2), \ldots, (x_n, y_n)\}$, which contains $n$ inputs of $d$-dimensionality, $x_i \in \mathbb{R}^d$, and their class labels $y_i$. In the case of binary classification, $y_i \in \{+1, -1\}$, where +1 and -1 are two classes of labels. In our study, $x_i$ represents the input parameters encompassing features characterizing both benign and malignant tumors in the Wisconsin Breast Cancer dataset, as well as molecular and clinical attributes pertaining to LGGs and HGGs in the TCGA dataset. In addition, $y_i$ denotes the classes in both datasets, such as benign and malignant for the breast cancer dataset and LGGs and HGGs for the glioma dataset. In SVM, a hyperplane defined by the normal vector $w$ and offset $b$ is determined based on the training data. Consequently, the following expression can be formulated for any data point $x_i$ and its associated class $y_i$.

$$y_i(w \cdot x_i + b) \geq 1 \quad (2)$$

The main objective of SVM is to find the maximum separation between two classes or the maximum margin ($2/\|w\|^2$) between them. In other words, it aims to determine a hyperplane that separates the two classes so that all points belonging to class -1 are on one side and all points belonging to class +1 are on the other. The hyperplane with the widest margin is referred to as the optimal separating hyperplane. The equation described above leads to a hard-margin SVM, which does not allow any points to lie within the margin, making it impossible to train the classifier, especially on linearly inseparable data [40]. Soft-margin SVMs, on the other hand, allow for some errors in classification by permitting some points to lie within the margin [40]. An appropriate separator is found while minimizing errors by allowing for some misclassification of data points. However, a soft-margin SVM can be introduced by allowing each data $x_i$ to deviate $\xi_i$ from satisfying the conditions in (2), obtaining a new set of conditions:

$$y_i(w \cdot x_i + b) \geq 1 - \xi_i \quad (3)$$

The variable defined here, $\xi_i$, is referred to as the slack variable and allows for a certain amount of misclassification. For an N-element nonseparable dataset, the dual optimization equation is given by:
maximize,

$$\sum_{i=1}^{N} \alpha_i - \frac{1}{2} \sum_{i,j=1}^{N} \alpha_i \alpha_j y_i y_j \langle x_i, x_j \rangle \quad (4)$$

subject to

$$\sum_i y_i \alpha_i = 0, 0 \leq \alpha_i \leq C \quad (5)$$

where $C \geq 0$ is a constant regularization parameter that minimizes the solution for which $\xi_i$ get larger. In this case, the decision function for classifying new data is expressed as follows:

$$f(x) = sgn\left(\sum_{i=0}^{N} y_i \alpha_i \langle x, x_i \rangle + b\right) \quad (6)$$

One of the most important features of SVMs is their ability to be applied to classification problems where classes are not linearly separable. A kernel function can be used to handle situations where the data are not linearly separable and to enhance the generalization capability of the SVM classifier [41]. In this case, a nonlinear mapping, $\Phi : X \mapsto F$, is used to map the input space ($X$) into a higher-dimensional space, called the feature space ($F$), and the data become linearly separable. Under this mapping of the decision rule, eq. 6 transforms into:

$$f(x) = sgn\left(\sum_{i=0}^{N} y_i \alpha_i K(x, x_i) + b\right) \quad (7)$$

where,

$$K(x, x_i) = \langle \Phi(x), \Phi(x_i) \rangle = \Phi(x)^T \Phi(x_i) \quad (8)$$

is the kernel ($K$) induced by the mapping $\Phi$.

A kernel is a type of data transformation function that helps transform data from the input space, where it is not linearly separable, into a higher-dimensional feature space. Additionally, the kernel contains the inner product in the feature space. However, given that the feature space can be extended to a very high-dimensional space, computational processing becomes difficult. In this case, the kernel trick is used to circumvent the problem of computational inefficiency in the extended feature space. Accordingly, there are many different kernel functions, such as polynomial, sigmoid and Gaussian radial basis functions (RBFs). However, although the SVM algorithm with kernels yields good results for classification problems, the evaluation of kernel inputs in a large feature space is computationally expensive. Moreover, as the feature space grows and the computational cost of estimating the kernel functions increases, there are limitations to the successful solution of such classification problems with the traditional SVM method.

*b) Quantum Kernel and Quantum Feature Maps*

In the classical SVM algorithm, the main motivation for using quantum kernels is to exploit the exponentially large target Hilbert space. This can lead to better data separability. Moreover, given the large size of the quantum Hilbert space, it offers many advantages, such as a large information capacity and noise resistance, which are especially useful for noisy medical data. Quantum computers can only operate on given data expressed as quantum states. In order to take quantum advantages, the classical medical data we use in this study need to be transformed into quantum states. Therefore, quantum feature mapping methods are used to nonlinearly map (encode or embed) classical data into quantum Hilbert space, a potentially much higher dimensional feature space [10]. Mathematically, a quantum feature map $\phi(x)$ is map from the classical feature vector $x$ to the quantum state $\langle \Phi(x)|\Phi(x)\rangle$, a vector in Hilbert space [7]. This is facilitated by applying the unitary state preparation circuit $U_\Phi(x)$ to the initial state, $|0>^{\otimes n}$, $U_\Phi(x)|0>^{\otimes n} = \Phi(x)$. Where $n$ represents the number of qubits used for encoding. Moreover, mapping two data inputs into a feature space and taking their inner product leads to a kernel function that measures the distance and similarity between data points. The quantum kernel entry for two feature vectors $x$ and $x_i$ is obtained as the Hilbert–Schmidt inner product, $K(x, x_i) = tr[\Phi^\dagger(x)\Phi(x_i)]$, and the quantum kernel matrix is defined as:

$$K(x, x_i) = \left|\langle \Phi^\dagger(x)\Phi(x_i)\rangle\right|^2 \quad (9)$$

and one can calculate the each element of this kernel matrix on a quantum computer by calculating the transion amplitude as:

$$\begin{aligned} K(x, x_i) &= \left|\langle \Phi^\dagger(x)\Phi(x_i)\rangle\right|^2 \\ &= \left|\langle 0^n|U_\Phi(x)^\dagger|U_\Phi(x_i)|0^n\rangle\right|^2. \end{aligned} \quad (10)$$

The kernel matrices computed from the quantum states of feature maps significantly influence the classification outcomes of the QSVM-Kernel algorithm. Therefore, selecting an appropriate feature map is fundamentally important for the classification performance of the QSVM-Kernel method. In this study, classical data were embedded in the quantum Hilbert space using nine different feature mapping methods, as specified below, and the effects of these mapping techniques on the classification performance of the QSVM-Kernel method were further investigated.

Angle encoding ($R_x(\theta)$, $R_y(\theta)$ and $R_z(\theta)$):

Angle encoding, or tensor product encoding, is a method of representing data values using the rotation angle of a qubit. The number of qubits depends on the amount of input data, and one qubit is required for each input vector component [10]. Moreover, this feature mapping method demonstrates notable efficiency in terms of time, as it requires only single qubit rotation gates for state preparation. In this study, three different single-qubit angle encoding techniques, namely, $R_x(\theta)$, $R_y(\theta)$ and $R_z(\theta)$ are employed.

$$R_x(\theta) = e^{-iX\frac{\theta}{2}} = \begin{bmatrix} \cos\left(\frac{\theta}{2}\right) & -\sin\left(\frac{\theta}{2}\right)i \\ -\sin\left(\frac{\theta}{2}\right)i & \cos\left(\frac{\theta}{2}\right) \end{bmatrix},$$

$$R_y(\theta) = e^{-iY\frac{\theta}{2}} = \begin{bmatrix} cos\left(\frac{\theta}{2}\right) & -sin\left(\frac{\theta}{2}\right) \\ sin\left(\frac{\theta}{2}\right) & cos\left(\frac{\theta}{2}\right) \end{bmatrix},\quad (11)$$

$$R_z(\theta) = e^{-iZ\frac{\theta}{2}} = \begin{bmatrix} e^{-i\frac{\theta}{2}} & 0 \\ 0 & e^{i\frac{\theta}{2}} \end{bmatrix}.$$

Amplitude encoding:

Amplitude encoding is a method of representing data using the amplitudes of a quantum state. In other words, data points are expressed based on the amplitudes of the quantum state [14]. In this method, $2^n$ classical feature vectors are encoded into the wave function of merely $n$ qubits. In general, the amplitudes of a quantum state can be encoded using the following equation.

$$|\psi_x> = \sum_{i=1}^{2^n} x_i |i> \quad (12)$$

where $|\psi> \in$ of the Hilber space and $\sum_{i=1}|x_i|^2 = 1$.

Pauli feature maps (ZFeatureMap, ZZFeatureMap):

The third mapping techniques used in this paper are ZFeatureMap and ZZFeatureMap from the Pauli feature map family. In particular, ZZFeatureMap is known to be difficult to simulate classically due to its entangling blocks. This is one of the key elements in gaining an advantage over classical approaches. The Pauli feature map ($U_{\varphi(\overline{x_i})}$) generated for $n$ qubits,

$$U_{\varphi(\overline{x_i})} = exp\left(i \sum_{S\subseteq[n]} \varphi_S(\vec{x}) \prod_{i\in S} P_i\right). \quad (13)$$

where $P_i$ denote the Pauli matrices and $P_i \in \{I, X, Y, Z\}$ and $S$ define the connectivity between different qubits [7, 16]. This feature map comprises layers of Hadamard gates interleaved with entangling blocks.

Parametrized quantum circuits ($R_x$ and CX, $R_y$ and CY, $R_z$ and CZ):

In this work, the last feature mapping technique used to encode classical data from classical vector space to quantum Hilbert space is 3 different parametrized quantum circuits consisting of single qubit and two qubit quantum gates. A parametrized quantum circuit is a unitary transformation that consists of a combination of parametrized (or non-parametrized) as well as fixed one- and two-qubit unitary quantum gates, which conduct operations on quantum states [19]. In general, parametrized quantum circuits are used in classification processes and have parameters (θ) that optimization processes can tune. However, since these circuits will encode the data from classical space to quantum Hilbert space in this work, these parameters will not be tunable and will only carry the input data. In this study, we used 3 different parametrized quantum circuits consisting of $R_x$, $R_y$ and $R_z$ single-qubit rotational gates to encode the data and CX, CY and CZ two-qubit quantum gates to generate entanglement between the data.

### III. RESULTS AND DISCUSSION

In this study, the QSVM-Kernel algorithm was employed for the classification of malignant and benign tumors in the Wisconsin Breast Cancer (original) dataset and for distinguishing LGGs and HGGs tumors in the TCGA Glioma dataset. Quantum simulations provided by the PennyLane 0.35.1 framework were utilized for computing special quantum kernel matrices formed by nine different quantum feature maps within the QSVM-Kernel algorithm. The quantum simulators provided by PennyLane are the default.qubit and lightning.qubit state-vector qubit simulators. In addition, leveraging the plugin framework, an external Python package providing additional quantum devices to PennyLane, the ibmq qasm simulator in the IBM Quantum Experience platform, was utilized alongside the other mentioned simulators. The pseudocode for the study is available in the **Supplementary Materials** section.

Following the determination of training and test quantum kernels, the training kernel was employed to train the Scikit-learn SVM model on a classical computer, whereas the test kernel was utilized to make predictions by transitioning to the model's prediction method. The classification performance results of the QSVM-Kernel model for both medical datasets are presented in Table 1.
The QSVM algorithms with ZFeatureMap and ZZFeatureMap were executed with a repetition of 2 using the lightning.qubit PennyLane simulator. On the other hand, QSVMs with other feature maps were run with a repetition of 1 using the default.qubit simulator. Moreover, in this study, to evaluate the classification performance of the QSVM-Kernel model, commonly used evaluation metrics in the literature were employed, including the area under the receiver operating characteristic curve (AUROC), accuracy, precision, recall, and F1 score measurements. The AUROC curve assists in assessing the balance between the true positive rate and the false positive rate. Other evaluation metrics were derived from the confusion matrix.

According to Table 1, when the QSVM-Kernel algorithm utilizes the $R_x$ rotational gate and the $R_x$ and CX parametrized quantum circuit as feature maps on the Wisconsin Breast Cancer (original) dataset, it achieves equal and optimal performance with accuracy and AUROC score values of 0.98 in distinguishing malignant and benign tumors. The total execution time for training and classification within the QSVM-Kernel algorithm is 19.25 min when employing the $R_x$ feature map and 44.1 min when utilizing $R_x$ and CX parametrized quantum circuits.
Additionally, when the $R_z$ rotational gate and the $R_z$ and CZ parametrized circuit are used as feature maps, they yield equal and worst classification results with an accuracy of 0.65 and AUROC score of 0.5.
The total execution times for these feature maps within the QSVM-Kernel algorithm are 15.38 min and 47.1 min, respectively.

**Table 1.** The computational and total execution time results of QSVM-Kernel algoritm for Wisconsin Breast Cancer (original) and TCGA Glioma datasets.

| Datasets | Feature Maps | Repetitions | Accuracy | AUROC | Total Execution Time |
|---|---|---|---|---|---|
| Wisconsin Breast Cancer (original) | $R_x$ | 1 | 0.98 | 0.98 | 19.25 min |
| Wisconsin Breast Cancer (original) | $R_y$ | 1 | 0.97 | 0.97 | 19.38 min |
| Wisconsin Breast Cancer (original) | $R_z$ | 1 | 0.65 | 0.5 | 15.38 min |
| Wisconsin Breast Cancer (original) | $R_x$ and CX | 1 | 0.98 | 0.98 | 44.1 min |
| Wisconsin Breast Cancer (original) | $R_y$ and CY | 1 | 0.97 | 0.97 | 46.63 min |
| Wisconsin Breast Cancer (original) | $R_z$ and CZ | 1 | 0.65 | 0.5 | 47.1 min |
| Wisconsin Breast Cancer (original) | Amplitude Encoding | 1 | 0.95 | 0.94 | 101.53 min |
| Wisconsin Breast Cancer (original) | ZFeatureMap | 2 | 0.94 | 0.94 | 41.22 min |
| Wisconsin Breast Cancer (original) | ZZFeatureMap | 2 | 0.96 | 0.96 | 42.01 min |
| TCGA Glioma | $R_x$ | 1 | 0.85 | 0.87 | 37.78 min |
| TCGA Glioma | $R_y$ | 1 | 0.86 | 0.87 | 38.83 min |
| TCGA Glioma | $R_z$ | 1 | 0.40 | 0.5 | 33.97 min |
| TCGA Glioma | $R_x$ and CX | 1 | 0.86 | 0.88 | 55.5 min |
| TCGA Glioma | $R_y$ and CY | 1 | 0.85 | 0.86 | 71.8 min |
| TCGA Glioma | $R_z$ and CZ | 1 | 0.40 | 0.5 | 82.62 min |
| TCGA Glioma | Amplitude Encoding | 1 | 0.84 | 0.86 | 152.15 min |
| TCGA Glioma | ZFeatureMap | 2 | 0.84 | 0.86 | 51.20 min |
| TCGA Glioma | ZZFeatureMap | 2 | 0.80 | 0.80 | 52.23 min |

Furthermore, considering all feature maps, the precision, recall, and F1 score values for the malignant cohort range from 0.40 to 0.95, 0.43 to 0.99, and 0.5 to 0.97, respectively, while for the benign cohort, they vary between 0.65 to 0.99, 0.97 to 1.0, and 0.78 to 0.98, respectively

Similarly, in the QSVM-Kernel algorithm, when kernels consisting of 9 different feature maps are used, the performance of this algorithm for discriminating between LGGs and HGGs is also presented in Table 1 under the TCGA Glioma dataset section.

According to the table, utilizing the $R_y$ rotational gate as a feature map in the QSVM-Kernel algorithm yields the best classification performance, with an accuracy value of 0.86 and an AUROC of 0.87. Following this, when the $R_x$ rotation gate is used as a feature map in the QSVM, an accuracy of 0.85 and an AUROC value of 0.87 are achieved.
Additionally, employing the $R_z$ rotational gate as a feature map results in the lowest classification performance, with an accuracy of 0.40 and an AUROC of 0.5.

The total execution times for the QSVM-Kernel algorithm with these three feature maps are 38.83 min, 37.78 min, and 33.97 min, respectively.
Moreover, considering the 9 feature maps, the precision, recall, and F1 score values for LGGs range from 0.60 to 0.97, 0.78 to 1.0, and 0.75 to 0.87, respectively. For HGGs, these values vary between 0.40 to 0.76, 0.91 to 1.0, and 0.57 to 0.85, respectively.

The confusion matrices of the model obtained when using the $R_x$ rotational gate as the feature map in the QSVM-Kernell algorithm on the Wisconsin Breast Cancer (original) dataset and when using the $R_y$ rotational gate as the feature map in the QSVM-Kernel algorithm on the TCGA Glioma dataset are illustrated in Figure 1.

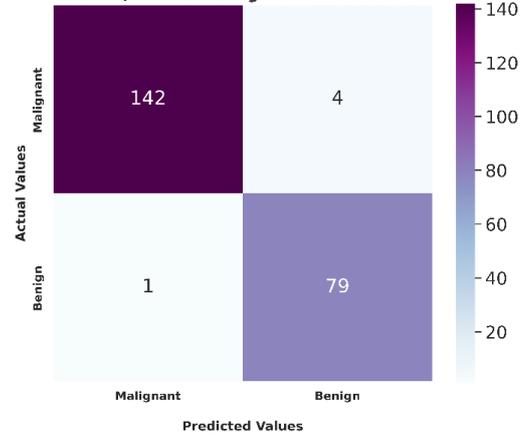

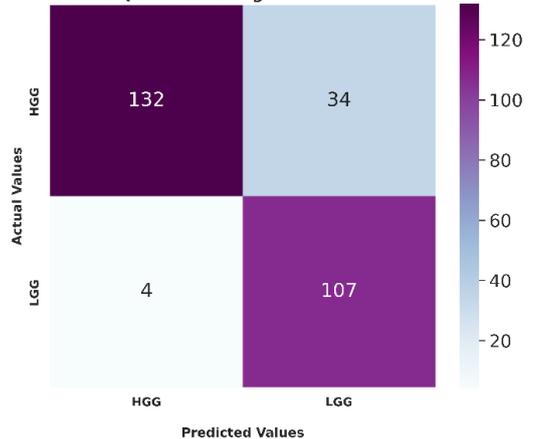

**Figure 1.** Confusion matrices of the QSVM-Kernel algorithm for the Wisconsin Breast Cancer (original) and TCGA Glioma datasets, in that order. These confusion matrices were obtained using the $R_x$ and $R_y$ rotational gates, respectively, are feature maps in each dataset's QSVM-Kernel algorithm.

## IV. FINDINGS AND IMPLICATIONS

The general findings obtained in this study are as follows:

- In datasets composed entirely of numeric data, such as the Wisconsin dataset, when mapping data from the classical vector space to the quantum Hilbert space, using the $R_x$ rotational gate yields better results in terms of both classification accuracy and execution time compared to other gates.
- For datasets consisting mostly of categorical data, such as the TCGA Glioma dataset, after converting data to numerical values through encoding, mapping from the classical vector space to the quantum Hilbert space using both the $R_y$ rotational gate and, in addition, the $R_x$ rotational gate proves to be more advantageous both in terms of classification accuracy and execution time.
- Using two-qubit quantum gates in a parametrized quantum circuit during the data mapping process significantly increases the total execution time, reducing efficiency.
- In medical data mapping processes, the utilization of $R_z$ rotational and CZ two-qubit quantum gates negatively affects both the classification outcome and execution time.

The results of the significant implications for the impact of different quantum kernels on the performance of the QSVM for classification, especially for complex medical data. The choice of quantum future maps significantly influences the outcome of classification and execution time of the QSVM algorithm. This study thus shows the importance of the right future maps in maximizing the accuracy and efficiency of QSVM-Kernel algorithms, especially in complex medical machine learning applications.

Therefore, this study has made the following two contributions:

1. shows the potential of quantum kernels for improving the performance of SVMs,
2. provide guidance for undertaking research and applications of quantum machine learning for improved classification performance, especially in medical data analysis.

## V. CONCLUSION

In recent years, with a significant increase in the volume of medical data, many classical machine learning models have been developed to expedite data analysis, manage datasets, and derive meaningful insights from these data. However, as the volume of data has grown, these advanced models have increasingly required more computational power, surpassing the capabilities of traditional computers equipped with CPUs and GPUs. As a result, quantum computing technology has been applied in the medical field despite the limitations of NISQ computers, leading to many successful outcomes. Revolutionary progress has been made in data analysis and problem solving in the healthcare domain with QML models, where the power of quantum computers is applied to machine learning tasks.

In this study, the QSVM-Kernel was applied as a QML model to classify 2 different and well-known Wisconsin Breast Cancer (original) and TCGA Glioma datasets. The QSVM-Kernel algorithm uses quantum kernel matrices generated by 9 different feature maps, namely, angle encoding ($R_x$, $R_y$, and $R_z$), amplitude encoding, Pauli feature maps (ZFeatureMap, ZZFeatureMap), and parametrized quantum circuits ($R_x$ and CX, $R_y$ and CY, $R_z$ and CZ). Thus, the classification performance of the QSVM-Kernel algorithm was analyzed according to the feature map chosen.

Considering the limitations of the NISQ computers in both datasets, the PCA value was set to 5. Accordingly, in the Wisconsin Breast Cancer (original) dataset, the QSVM Kernel algorithm achieves equal classification performance when $R_x$ rotational gate and $R_x$ and CX parametrized quantum circuit are used as feature map to distinguish malignant and benign tumors. However, since the total execution time is about 2 times higher for the parametrized quantum circuit ($R_x$ and CX) and to ensure energy efficiency in quantum circuits, it is more advantageous to use only the $R_x$ rotational gate. Moreover, in the TCGA Glioma dataset, the best classification performance was obtained when the $R_y$ rotational gate was used as a feature map in the QSVM-Kernel algorithm. In this dataset, although the $R_z$ rotational gate gives better results in terms of total processing time, the classification performance value is too low to be generalized when it is used as a feature map.

In the future, we aim to develop efficient feature mapping techniques, especially for medical datasets, that will positively affect both the classification result and execution time and use them in the QSVM-Kernel algorithm.

**Conflict of Interest**

The authors declare that they have no conflicts of interest.

**Data Availability**

The datasets used in this study are publicly available. The Breast Cancer Wisconsin (Original) dataset can be accessed from the UCI Machine Learning Repository at the following link: Wolberg, W. (1992). Breast Cancer Wisconsin (Original). UCI Machine Learning Repository.
The Cancer Genome Atlas Program (TCGA) dataset is available from the National Cancer Institute at the following link: The Cancer Genome Atlas Program (TCGA) - NCI.

**Ethical Statement**

The datasets used in this study are publicly available and were obtained from the UCI Machine Learning Repository and the National Cancer Institute's Cancer Genome Atlas Program. Both datasets are de-identified and do not contain any personally identifiable information. Therefore, our study did not necessitate approval from an institutional review board or adherence to the Health Insurance Portability and Accountability Act.

**Funding Statement**

This work was supported by the Research Fund of Yildiz Technical University [Project Number: 5936].

**Acknowledgments**

The authors would like to acknowledge that this paper is submitted in partial fulfillment of the requirements for PhD degree at Yildiz Technical University.

**Supplementary Materials**

Pseudocode for the QSVM-Kernel Algorithm: This pseudocode outlines the QSVM-Kernel algorithm, including the steps for data preprocessing, quantum circuit definitions, and classification processes, incorporating various quantum feature maps.

**Supplementary Materials**

The figure illustrates the complete workflow, including data preprocessing, quantum circuit definitions, and classification steps, using various quantum feature maps. It encompasses the processes for quantum embeddings, rotational gates, parametrized quantum circuits, amplitude embeddings, and ZZ and ZFeatureMaps, followed by the training and evaluation of the SVM model.

---

**Pseudocode for the QSVM-Kernel Algorithm**

---

```
// Import necessary libraries and modules
import Pennylane
import Qiskit
import Sklearn
import Pandas
// Import other necessary packages

// Load and preprocess the data
Read data from "(dataset).xlsx"
Extract and preprocess labels and features
Split data into training and test sets
Apply PCA, StandardScaler, and MinMaxScaler to scale data

// Set up quantum device
Initialize quantum device with n_qubits and shots=1000

// Define function for quantum embeddings and SVM classification
Function quantum_embedding_and_svm(rotation_type, training_data, training_labels, test_data, test_labels):
Define quantum circuit for angle embedding with rotation_type
Define kernel function using quantum circuit
Train SVM with quantum kernel
Predict and evaluate test data
// Perform quantum embeddings and SVM classification for RX, RY, RZ rotations using a loop
for rotation in ["X", "Y", "Z"]:
Call quantum_embedding_and_svm(rotation, training_data, training_labels, test_data, test_labels)

// Define function for parametrized quantum circuits with controlled gates
Function parametrized_quantum_circuit(rotation_type, controlled_gate, training_data, training_labels, test_data, test_labels):
Define quantum circuit for angle embedding with rotation_type and controlled gates
Define kernel function using quantum circuit
Train SVM with quantum kernel
Predict and evaluate test data
// Perform parametrized quantum circuits using a loop
For (rotation, gate) in [("X", "CNOT"), ("Y", "CY"), ("Z", "CZ")]:
Call parametrized_quantum_circuit(rotation, gate, training_data, training_labels, test_data, test_labels)

// Define and apply amplitude embedding
Function amplitude_embedding(training_data, training_labels, test_data, test_labels):
Define quantum circuit for amplitude embedding
Define kernel function using amplitude embedding
Train SVM with quantum kernel
Predict and evaluate test data
Call amplitude_embedding(training_data, training_labels, test_data, test_labels)

// Define and apply ZZFeatureMap for ZZ kernel
Function ZZFeatureMap(n_qubits, data):
Apply Hadamard gates to qubits
Apply RZ gates with data
Function apply_ZZ_kernel(training_data, training_labels, test_data, test_labels):
Define ZZ kernel function using ZZFeatureMap
Train SVM with ZZ kernel
Predict and evaluate test data
Call apply_ZZ_kernel(training_data, training_labels, test_data, test_labels)

// Define and apply ZFeatureMap for Z kernel
Function ZFeatureMap(n_qubits, data):
Apply Hadamard gates to qubits
Apply RZ gates with data
Function apply_Z_kernel(training_data, training_labels, test_data, test_labels):
Define Z kernel function using ZFeatureMap
Train SVM with Z kernel
Predict and evaluate test data
Call apply_Z_kernel(training_data, training_labels, test_data, test_labels)
```